# New Ideas for Brain Modelling 5

Kieran Greer, Distributed Computing Systems, Belfast, UK.
http://distributedcomputingsystems.co.uk
Version 1.5

***Abstract***—This paper describes a process for combining patterns and features, to guide a search process and make predictions. It is based on the functionality that a human brain might have, which is a highly distributed network of simple neuronal components that can apply some level of matching and cross-referencing over retrieved patterns. The process uses memory in a dynamic way and it is directed through the pattern matching. The paper firstly describes the mechanisms for neuronal search, memory and prediction. The paper then presents a formal language for defining cognitive processes, that is, pattern-based sequences and transitions. The language can define an outer framework for concept sets that are linked to perform the cognitive act. The language also has a mathematical basis, allowing for the rule construction to be consistent. Now, both static memory and dynamic process hierarchies can be built as tree structures. The new information can also be used to further integrate the cognitive model and the ensemble-hierarchy structure becomes an essential part. A theory about linking can suggest that nodes in different regions link together when generally they represent the same thing.

## 1 Introduction

The human brain is known to store a vast amount of information through the many links it creates between its neuronal components. This information needs to be found again and manipulated to match with any current situation. Because the real world is even more complex, it is not possible to imagine that a brain would store a static result for every scenario. Instead, it must be able to make use of its memories in a more dynamic way, where shallow hierarchies can provide a useful 'unit of work' that would be more complex than individual neurons. When put together at a larger scale, repeating structures of this type can often provide a much greater degree of complexity and this paper considers how that might be applied to a brain model.

This paper firstly considers how memory can be found and joined, to make real concepts and even form predictions. This paper then presents a more formal approach to building cognitive models using the same brain-like structures. The purpose of the formal approach is to be able to define a scenario that can be consistently translated over to a pattern-based model and used as a starting point for distributed learning algorithms. A pattern of nodes can represent something and in that pattern, there are sub-patterns that represent the sub-concepts or features of the outer pattern. Through feature-matching, patterns and sub-patterns can be linked and used to retrieve relevant structures from memory, to guide a search process. If the process is carried out at the coarser symbolic level, it is much more generic and can act as a framework for the neuronal architecture. This generic structure is used by the cognitive model and is developed further with some new ideas.

The architecture of concern is described in Figure 1, developed later in Figure 2, with a suggestion related to biology being given in Figure 3. Figure 1 relates to the process specification of section 5 firstly, but more specifically the updated version in section 6. It demonstrates how pattern ensembles can translate over to a more structured hierarchy and more importantly, how the process language rules can be integrated into the model. By using an appropriate language, tree links can create process cycles that add direction to the ensemble transitions and emerge largely from the linking structure itself. The figure also shows some normalisation through redundancy, which is what a database might perform and a real brain might perform slightly less. Therefore, the design of Figure 1 may be more exactly what a computer would use, whereas a brain-like structure would prefer the designs from section 6 onwards.

The rest of the paper is organised as follows: section 2 describes some related work. Section 3 describes a new theory for systematic search and prediction, while section 4 finishes the memory discussion with some pattern and hierarchy relations. Section 5 introduces a formal script for defining cognitive processes and an appropriate test paradigm. Section 6 improves the script through a cognitive language that has a mathematical basis. Section 7 then puts the script in terms of the current cognitive model, while section 8 gives some conclusions on the work.

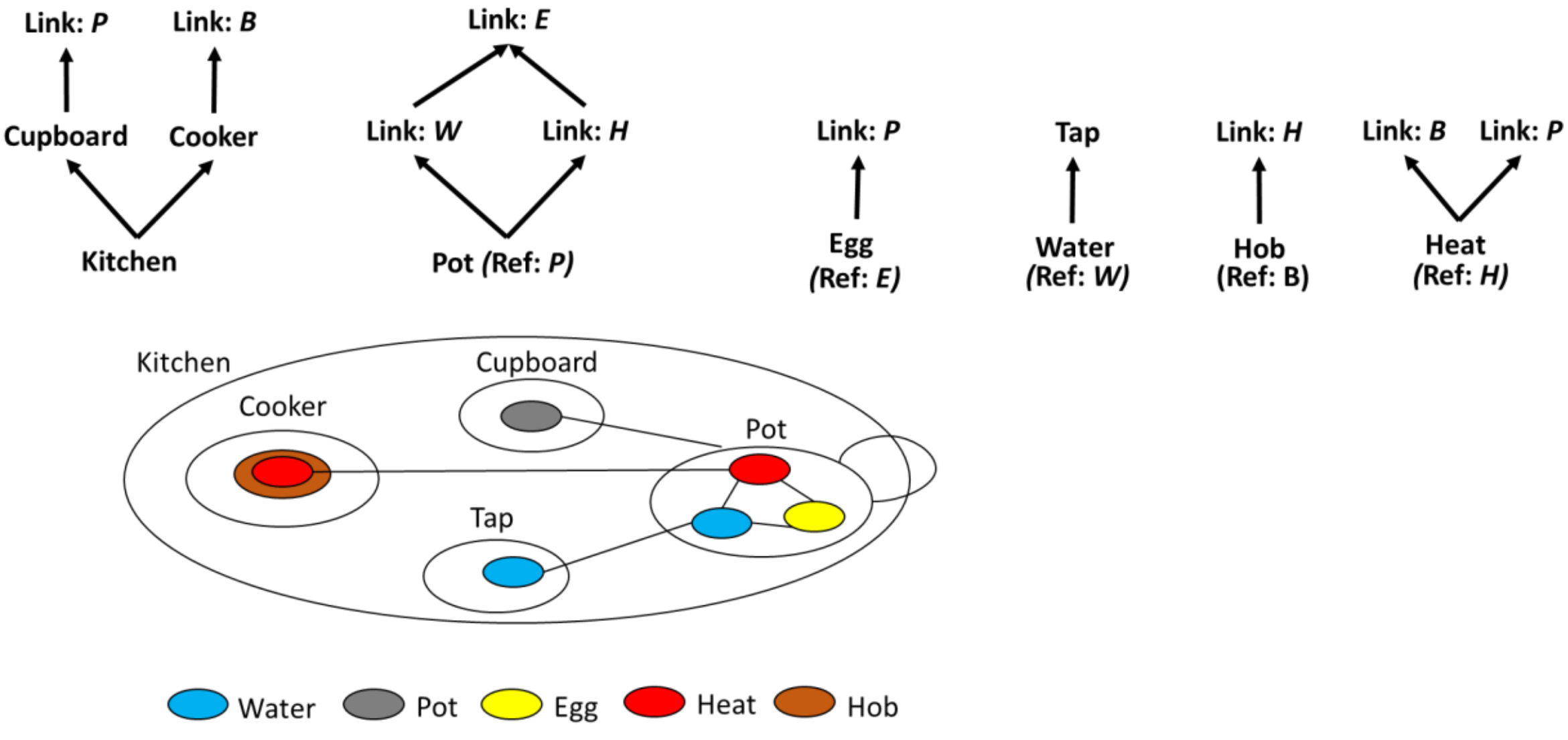

Figure 1. The Pattern Ensemble below the Concept Trees can synchronise their activation.

## 2 Related Work

### 2.1 Related Work from AI

The author has described his own model in earlier papers [8]-[14], which represents a cognitive model [13] that spans linking nodes into patterns, to intelligent processing. The idea of a fully-connected neuron structure is quite common (for example, [1][17] and some of the author's work). Image recognition is the primary input sense and was introduced, along with behaviours, in [10]. Considering earlier papers again, a tree-like structure known as a Concept Tree was defined in [12]. Linking patterns can be done at the level of the ensemble, but another option is to use a hierarchical structure and so pattern ensembles can translate over to tree-like structures. If hierarchies exist, they can still combine cross-referenced regions or shallow hierarchies in a next level, similar to convolutional deep learning [18].

A new process language is suggested, but there already exists formal languages for this. Examples would be modelling Business Processes using BPEL [3] or BPML [25]. Simulations can use something like Specification and Description Language (SDL) [23] and Agent-based languages are also well known about, such as the FIPA Agent Communication Language [6]. A Markov Decision Process (MDP) [2] can define state transitions with choices, probabilities and rewards. The new 'Cognitive Process Language' suggested in section 6, has actually been finalised in a second paper [8], but the original ideas for it are described in this paper first. It is really only an outline but is closely linked with the structure that it creates. At this stage, the language also looks like an Ontology, with concepts and links between them.

## 2.2 Related Work from Biology

### 2.2.1 Neocortex Region

The paper [15] is an early attempt at modelling the neuron, or the cortex in particular, more exactly. The author thinks that there are some interesting overlaps with his own model. For example, gating and adaptive tuning is where a neuron or neuron circuit is prepared for firing, something also written about in [17]. The biological design in [15] however has features such as a triple relation of position, movement and target, between modules. This starts to look like the process language described later. While it does not match with the language exactly, there is also a structure known as a synaptic triad [5]. With that, the efficacy of a synapse of neuron A on neuron B can be influenced by the activity of a third neuron C, called a modulator. Timing and adaptive timing is thought to relate to bistable states and a greater force being able to reactivate something (nested patterns in [11]). The quote 'Receptive fields of associative units are "photographies" of the pattern' suggests abstraction, from image parts to the memory prototype through a hierarchy (as in [9]), and there is also an 'and-or' architecture (as in [10]). They state that:

> 'The appeal of the neocortex relies on the following facts: i) greater regularity is observed in its organisation; ii) cortical circuits process any set of data with a single format, whatever the modality; and iii) the phylogenetic development of the cortex is correlated with the acquisition of higher level behaviour structures.'

However, the subcortical systems also need to be included in the same single framework. If considering higher-level reasoning, then the cortex region has also been modelled in [16][17]. Their design also models the columnar structure of the cortex. As part of their prediction theory, a neuron can again be depolarised slightly, which involves the idea of a weakened signal preparing neurons for firing. The main idea is that the neuron can learn through active dendrites, different patterns for activation from sparse coding in the input synapses. This idea of plasticity is also written about in [27]. A biological description of the formation and storage of memories can be found in [20]. Cross-referenced memories would also set up preferred regions for activation, if they find or vote for features that remain persistent during the search.

#### 2.2.2 Linking Ensembles and Hierarchies

An ensemble-hierarchy relationship is implicit in the work of this paper and there may be some tentative associations with real brain modelling. The paper [4] describes tests on the cortex region and two types of excitatory population – specific and non-selective. The specific neurons make a categorical decision by recognising a signal type. The non-selective neurons would be more like an ensemble, but are still part of the decision-making process, as shown in their figure 13. All neurons are fully-connected and are studied using reduced mean-field equations. Stable attractors of interest are created, from the fully-connected nature of the network. The paper [21] shows binary connections between pattern ensembles, but often of the same type/brain region. As well as ensembles, they also found linear clusters of regions grouped according to neuroanatomical criteria – due to spatial closeness or homofunctional spacialization (hippocampus, cerebellum and motor control, including cognitive processes). They report that this also shows a remarkable construction of the hierarchy from a random state, done through link strength alone. Looking at Figure 1 again and as described in section 6 onwards, the linking density and related firing frequency in the figure might be an indication of the direction of flow in the hierarchy. If that is the case, then the ‘Pot’ concept should be at the bottom, followed by its sub-concepts of ‘Water’, ‘Heat’ and ‘Egg’, with the larger but less-used concepts of ‘Cupboard’, ‘Tap’ and ‘Cooker’ at the periphery. It is not impossible to see this type of structure in the real brain modelling of [21] (figure 6), with the thresholds set to 0.5 or 0.52, for example. Pyramidal neurons can be suggested as part of the design, with multiple links to the base and a more structured output. If the structures also switch roles in different patten sets, then keeping them distinct is another problem.

### 2.2.3 Modular Work Units

The Small World property [22][26] has been used to describe modules in the brain that are compact units, larger than a cortical column, for example. The paper [22] gives a review of why brain networks should be modular, which includes module hierarchies and sub-modules. The paper notes both anatomical and functional modular networks and states that: 'Hierarchical modularity specifically also enhances dynamical reconnectability (Robinson et al.), as marginally stable networks can be combined or divided while preserving stability.' Also: 'Models with adaptive rewiring, such as coupled maps with variable coupling strength (Rubinov et al.), typically incorporate a reinforcement of links between synchronized units and a pruning of links between asynchronized ones. This feedback between structure and dynamics, similar to synaptic plasticity in neuronal dynamics, naturally drives the emergence of inhomogeneities and modules in networks.' The cited Rubinov et. al. paper [24] supports these theories and explains that random structural connectivity guided by emergent synchrony patterns has been shown to evolve to modular small-world connectivity [7]. Functional modules can be coupled with apparently random structural ones. They state: 'The absence of modularity in a random network renders it unlike the known connectivity of the cortex. However, the spatiotemporal activity that unfolds on this structure evidences partial synchronization amongst the weakly and randomly coupled nodes, resulting in a modular functional connectivity matrix.' Also: 'We hence seek a detailed exploration of the nature of this structural self-organization. We observe that, as in (their) Figure 1, coupled chaotic dynamics generate ordered, modular functional patterns, even on random structural networks. ... We find that central hub nodes play a key role in the cohesiveness of this small-world network. … The model hence represents a crude approximation of Hebbian learning in a spontaneously active – or "resting state" – ensemble of coupled oscillators.' Although, the modularity is in the functional network that is created and not the random structural one; and new neurons are not created but existing ones are re-wired. This synchronization is sometimes written about as quantum, or regions firing at the same time.

## 3 A Theory for Search and Prediction

This section describes the first theory in the paper. It is intended to model one aspect of the pattern-matching process of a human brain. The brain's neocortex region [17] is responsible for the higher-level or intelligent functions, but silent searches need to find and filter the stored memories first. A number of different elements are involved. Firstly, there is the sensory input. This would determine where the brain starts from and would be a persistent input signal. Then there are also the memories that the brain has stored from earlier, similar experiences. To reason over this input, some type of search is required, to move through pattern phases from the input to a desired final state. This involves remembering actions, scenes or events from the past and applying them to the present, as a type of prediction. Recognising a pattern is also the task of recognising relevant features in the pattern. How this may work at a neuron-level is considered in other papers and so this paper assumes that any input pattern, feature set, or related memory can be found as required. One question might be if this would be a difficult process for the brain to realise through a neuronal component, but it may simply be the result of more connections guiding a search process. The following sub-sections consider how these elements can be manipulated, to produce some level of reasoning and finally predict some type of event from that reasoning.

### 3.1 Static and Dynamic Memory

If we receive input from our environment, the easiest approach would be to store an exact result for each situation encountered, but this would probably require too many static memories to be practical. Another idea might be to cross-reference shared memories so that they can be used more dynamically and probably even in previously unencountered scenarios. The cross-references can also be used to guide a search, which would be constrained by the linking structure. As such, there is no requirement for a fixed structure for every scenario, but mix and match can be tried until something useful is found. The memory and pattern matching are therefore defined by more densely connected regions. Specific features can be recognised through nested regions that are again more consistently connected. If a pattern is fully-connected, then the size reduction can be quadratic, or the connection size could be reduced from the square to the number of nodes only. This allows the process to scale in a small number of levels to a much smaller size, which is helpful for aggregating or abstracting. This fully-connected structure is therefore quite basic and might be looked at as a compact unit of work.

### 3.2 Feature Selection through Pattern Cross-Referencing

Therefore, the first phase is to retrieve the related memories. There would be an input sensory signal and it is assumed that the brain can parse this and recognise the relevant features. If a feature is recognised, then this should trigger a memory process that tries to retrieve previous scenes that also contained the feature. For example, one can imagine simply a visual input of a room with items, where each item can also be remembered from a past experience. Each item can have its own links to related scenes and experiences. Therefore, if we consider the brain to have parsed the image into a list of features, then each feature has retrieved another similar list. Therefore, to summarise the current situation, it is only required to aggregate over the returned feature lists, to get from memory what the current scene is mostly about. This is simply a cross-referencing process and it could even be a majority vote, where the reduced feature set would result in the memory's understanding of the situation. Any reasoning must start with the input signal pattern and so it is a matter of then restricting that with the memory's understanding, to try to predict what the best action might be.

### 3.3 Realising Predictions

A prediction could therefore be seen as a constraint on the situation. There may be many choices, but only a smaller number of them will be any good and the prediction is concerned with guessing what the best option might be. It is also constrained by what is legal in the current situation, where a memory-related search cannot go somewhere if it is impossible in the current scene and the scene is unlikely to go against the pattern mass returned by the memory. A feature may also represent a future event that is not in the current scene, but that has been important in the memory. This would help to guide the search and could indicate a completed cycle, if it eventually matches with a current scene. The prediction therefore matches the scene features with the memory features, that may create a new scene to continue the search from, and that should eventually complete a cycle.

This can all happen using shallow hierarchies – two to four levels even, but it would be very expensive with regard to the number of nodes and connections, and the size when scaling-up would increase quadratically. As with modern neural networks, the input scene can be broken up into regions and the process applied to each, when the quadratic expansion is not as bad. If those aggregated regions are then aggregated again, a general plan or prediction for the whole scene can be achieved. It is probably unlikely that every stage of the search process would produce something significant. It is more likely that terminal states will get recognised and those results will be more significant. Therefore, a lot of the search can be silent, where the processes do not register with the conscious.

## 4 Patterns in Hierarchies

As described at the end of section 3.3, memories would like some type of terminal point, to allow them to be distinguished from a transient search stage. It may be thought that this would require aggregations through depth to some end point, but in fact the search can use shallow hierarchies, even for complex concepts and cycling can also indicate end points. If the shallow hierarchy is a unit of work however and linked with other ones, then it is maybe more a question of semantics, about where a concept or hierarchy starts or ends.

Central to this then is the idea of feedback from an image to itself and a human may not generally search outside of this constraining domain, represented only by the skills and knowledge of the agent in question. Any reasoning process would also have very clearly defined mappings between state changes with results, that would allow a search process to consider mostly only those mappings. In fact, this event-result pairing may be part of the heuristic from the beginning and may determine how the heuristic develops. For example, Autonomic Computing [19], based on the human nervous system, uses this type of control loop in the form of sensors and effectors. The sensors and effectors provide input to the system that may require some type of reciprocal response. Humans therefore use heuristics in the same way as a computer, applying previous knowledge and judgement to estimate some answer. What is different in humans is the fact that they can evolve the heuristics and so the heuristic can become a more dynamic process. Hierarchies can represent functional changes as well as memory ones, for example. In earlier work [10], this learning process spanned from images to behaviours, that update themselves through feedback and self-recognition, and may have a strange object-oriented nature to it, where memory images are linked with processes performed on them. No two people see the world in the same way.

### 4.1 Concept Condensing

As so much structure is required, it is important to make it as economic as possible. If cross-referencing is used, then static memory objects can be used in different scenarios. One option would be to store relative information such as location or position in a different brain region, that links to abstract concept representations only. If the object was represented in two different places, for example, it would it be uneconomic to store the whole memory description twice and so a reduced representation could be repeated instead. Intermediate neurons, for example, can link back to the original memory objects, but still have complex firing networks in their own right.

## 5 Formally Defining Pattern Sequences

This section starts the second half of the paper and presents a more formal notation for describing a scene that the distributed model would try to learn. The script is only able to define the outer framework for the model, where a complete solution would require other configurations, such as numbers of nodes in patterns, learning algorithms, and so on. It is mostly set-based, but it is also able to define relations between the patterns. If the outer framework can be built without ambiguity, it can then be easily repeated and understood. This section provides a foundational introduction, but is otherwise mostly superfluous. Section 6 then refines the process language, using a mathematical basis and shows how this first attempt may have some errors. Note that the paper [8] then makes further corrections and gives examples of the language use.

### 5.1 An Example of a Problem and Process

This section creates a script for the problem of boiling an egg. For this problem, the concepts include the kitchen, cooker, sink, pot, water and the egg. The kitchen is really the outer-most concept that covers everything else, but a set of instructions for the problem could look like the following:

1. Go into the kitchen and get the pot.
2. Go to the tap in the kitchen and fill the pot with water.
3. Go to the cooker in the kitchen and switch the heat on.
4. Place the egg in the pot and the pot on the cooker heat and wait for it to boil.
5. Monitor the heat level.
6. Wait for the egg to cook.

This can be a straight sequence of patterns, representing the instructions or acts. At some point however, the pattern may repeat. For example, if the water gets too low, then go to the tap again and add some more water. Or if the cooker temperature gets too high or low, then adjust it. So, there could be a backwards link from point 6 to point 5, or points 5 or 6 to point 2, for example.

For the script to be generic, variable actions are not allowed as entities, when the static object clusters are all that can be considered. Therefore, the instruction set should be re-written based on the object clusters only, where the reduced description would look like:

1. Kitchen -> Pot.
2. Kitchen -> (Tap -> Water) ∧ (Pot -> Water).
3. Kitchen -> (Cooker -> Heat).
4. Kitchen -> (Pot -> Water) ∧ (Pot -> Egg).

Converting this into object sets and relationships could result in the following:

Kitchen(Tap(Water), Cooker(Heat), Pot(Water, Heat, Egg)).

Tap -> Cooker -> Pot.

In this description, 'heat' and 'heat source' may be the same thing. Should sensory input be modelled as an object if it has a material influence? With processes, we are maybe more concerned with these objects of influence, whereas with memories, we are maybe more concerned with static structure. The final instruction describes the state transitions, from tap to cooker to pot. It might also be assumed that the same (sub)concept can be automatically backwards linked to any matching one, as a type of feedback.

### 5.2 Test Paradigm

Even though the pattern sequences are defined, the distributed system is still required to create and link the nodes inside of each pattern, to see how they evolve and interact. A problem like selecting the best learning mechanism can then be solved. It would therefore be possible to start with random node sets inside of the framework and try to evolve them, to match with the outer framework in some way. The more traditional method would be to present sets of nodes to the system and ask it to correctly learn their links. If also learning pattern sequences, then rules would probably be added and learned, maybe from a pre-defined script. The difference would be that it may only consider small bits of the problem at a time and the method cannot learn beyond what it has seen. If the goal is known, then the learning method can try to evolve into that goal, which is at least another option. It may be more organic, or suit a resource-constrained environment better.

## 6 A Cognitive Process Language

The earlier sections have given a theory for a process language and put it in the context of the current cognitive model. Section 5 describes a first attempt at the language, but in fact has some errors. This section extends the idea with a formal mathematical basis that will also increase the language's verbosity and power. The second attempt is being called a Cognitive Process Language (CPL) and the mathematical basis for it surprisingly has relations with Newtonian Mechanics, although the related work section describes how triple relations do occur in the real brain structure. While the first attempt uses the method of associating related concepts in two's, the new language requires at least three entities for each statement. These entities can be called E1, E2 and E3. Two of the entities are already associated and could include nesting, but would remain in their current state unless influenced by the third entity. The third entity is therefore external to the other two and requires the influence of only one of them for its own state change.

Consider a scenario where entities E2 and E3 are already associated and would remain as is, except that entity E1 is a new influence. E1 requires E3 to change its own state and so it is the effect of E3 that binds the rule. For tree-like structures, nesting can be defined, where in this case E3 is a sub-concept of E2 and after the rule, E1 is associated with E3 but is still separate from it. It should be possible to trace this type of rule through every statement that is declared as part of a scene and every statement should be consistent with every other statement in that scene. With 3 entities, the rule is of the form:

$E1 + E2 \cdot E3x \rightarrow E1 \cdot E3y + E2 \cdot E3x\text{-}y$, where $E3 < E2$ and $E1 - E3$, for example.

The hypothesis is based on the observation that we use this type of triple when interacting with the real world. Newtonian Mechanics states that a force stays as it is unless acted upon by another force, so that is the basis for the theory. Therefore, this third entity can bind the associations to a consistent sequence that can be mathematically worked out. The input of E3 to E1, also represented in the RHS description, gives the rule some direction. For example, consider that E1 represents a larger tree and is looking to combine with E3, and so it may have a receptor node for the type E3, but E3 can only be measured through the parent entity E2. E2 knows what self is, but E1 does not and so when combining with this smaller tree, E1 needs to know both about E3 and E2. E1 wants to link with E3 directly, but it also needs the reference to the E3 source and so the E2.E3 structure can be inverted first and then linked with E1. This gives a new linked path in the larger tree, or if part of that path exists, the rest is maybe added to it. The specification from section 5.1 can now be updated using the new triple rules and it is shown that this will help to remove errors in the coding. The description can now be simplified to the form:

$E1 + E2 \cdot E3 \rightarrow E1 \cdot E3 \cdot E2$.

If E3 is available from another source E4 as well, then the E3 branch can have two sub-branches to E2 or E4, for example. As is typical with process descriptions, multiple inputs or outputs can be described as follows:

E1 + E2·E3 Λ E4·E5 → E1·E3·E2 Λ E1·E5·E4,     for multiple inputs, or
E1 Λ E4 + E2·E3 → E1·E3·E2 Λ E4·E3·E2,     for multiple outputs.

Nesting or link relations can therefore be declared for each rule, based on the associations, and define a tree ordering that should be maintained as much as possible. In the script, a repeat process can be declared as a self-loop on a concept. In that case the concept or any of its sub-concepts can repeat themselves. The self-references may also be specified as associations, for example: E1 → E1. What is important is that one rule does not break any associations in any other rule for the same scene. The syntax for the language would therefore be as described in Table 1.

| E2·E3 | Entity E2 gives effector Entity E3 |
|---|---|
| E1 + E2·E3 | Entity E1 requires input from Entity E3 provided by Entity E2 |
| E1 + E2 + E3 | Entities E1, E2 and E3 all require input from each other in some way. |
| → E1·E3·E2 | Results in Entity E1 being affected by Entity E3 from source Entity E2 |
| E3 < E2 | Entity E3 is a sub-concept of Entity E2 |
| E3 – E2 | Entity E3 is associated with Entity E2 but separate from it |
| E2·E3 Λ E4·E5 | The input requires two effector entities, provided by two others |
| E1 Λ E4 | The effector entity provides output to two other entities |
| Self-reference | The concept loops on itself, including any sub-concept |

Table 1. Cognitive Process Language Syntax

### 6.1 CPL Cooking Example

The problem of cooking an egg might now contain the following entities: kitchen (K), cupboard (D), cooker (C), hob (B), gas (G), heat source (H), pot (P), water (W), egg (E) and tap (T). The instruction sequence can be the same and is repeated below:

1. Go into the kitchen and get the pot.
2. Go to the tap in the kitchen and fill the pot with water.
3. Go to the cooker in the kitchen and switch the heat on.
4. Place the egg in the pot and the pot on the cooker heat and wait for it to boil.
5. Monitor the heat level.
6. Wait for the egg to cook.

The first step to retrieve the pot could be:

$P + K \cdot D \rightarrow P \cdot D \cdot K$, where $D < K$ and $P - D$.

While the pot starts inside the cupboard, it is not really a sub-concept of it and while the cupboard is the variable that changes, there is only an association afterwards. It may help to think that the pot is never used while in the cupboard. The rule to fill the pot with water could be:

$P + T \cdot W \rightarrow P \cdot W \cdot T$, where $W < T$ and $W < P$.

So the change action is the water moving from the tap to the pot and it ends up being a pot sub-concept as well. Then the cooker needs to be turned on, which means to ignite the heat source, as in:

$H + C \cdot B \rightarrow H \cdot B \cdot C$, where $B < C$ and $H < B$.

If heat and heat source are modelled as the same thing, then this rule can be tried. More accurate might be to add the gas concept as follows:

$H + C \cdot B \cdot G \rightarrow H \cdot G \cdot B \cdot C$, where $B < C$ and $H < G < B$.

In this case, gas is the variable to be measured and is linked to the cooker through the hob concept. This is an example of some flexibility that is possible. Placing the egg in the pot is actually to place it in the water and not directly in the pot, as in:

$E + P \cdot W \rightarrow E \cdot W \cdot P$, where $E - W$.

The earlier rule states that the water is contained in the pot but it is difficult to state that egg is a sub-concept of water, so that can be made an association. Then to cook the egg requires the pot to be placed on the hob, when the heat then influences the egg. This can describe this as follows:

$P + B \cdot H \rightarrow P \cdot H \cdot B$, where $P - B, H < P$.

$E + P \cdot H \rightarrow E \cdot H \cdot P$, where $E - H < P$.

It is probably not correct to specify that pot is a sub-concept of hob, so that can be modelled as an association. To monitor and cook the egg would mean to go back from the water heat to the hob heat. It would also require a self-loop on the pot, which would allow a repeat of the hob interaction, where the final two rules could therefore be as follows:

$B + P \cdot H \rightarrow B \cdot H \cdot P$.

$P \rightarrow P$.

The first final rule however is a reverse of an earlier rule and both simply indicate a process that needs to be repeated. So that is what has to be modelled in the tree. This would lead to the following set of rules to describe the cooking process:

1. $P + K \cdot D \rightarrow P \cdot D \cdot K$, where $D < K$ and $D - P$.
2. $P + T \cdot W \rightarrow P \cdot W \cdot T$, where $W < T$ and $W < P$ and $P - T$.
3. $H + C \cdot B \rightarrow H \cdot B \cdot C$, where $B < C$ and $H < B$.
4. $E + P \cdot W \rightarrow E \cdot W \cdot P$, where $E - W$.
5. $P + B \cdot H \rightarrow P \cdot H \cdot B$, where $P - B, H < P$.
6. $E + P \cdot H \rightarrow E \cdot H \cdot P$, where $E - H < P$.
7. $B + P \cdot H \rightarrow B \cdot H \cdot P$. (not new, indicates a repeat for process 5)
8. $P \rightarrow P$.

With any human activity there can be increasing complexity with knowledge, which would be difficult to represent through a linear set of concepts. Therefore, any statement at one level in the language should be allowed to store any number of sub-statements that describe some aspect of its domain in more detail. For example, more precise details about how to cook the egg could be added to the 'Pot' rule-set. Finally, converting this into object sets and relationships could result in the following:

Kitchen(Cupboard(Pot), Cooker(Hob(Heat)), Pot(Water, Egg, Heat), Tap(Water)).

This would translate over to the ensemble patterns very easily and the pot and water links are already clear. It is also clear that 'Pot' is the most linked concept and therefore also the most frequently used. Other links would then be derived from it. The earlier Figure 1 shows that it is possible to display these rules graphically, while the next section adds some direction to them through naturally occurring cycles.

## 7 Abstracting the Formal Description to the Cognitive Model

The instructions of section 6.1 now look more like set tuples or horn clauses, which gives some match with the concept database of [14]. The ensemble structure is in the lower half, while the hierarchy network is in the upper half. This also complies with the structure of Figure 1. The ensemble is largely undirected, while the tree links can give direction in the upper structure, as described next.

### 7.1 The Cognitive Model

If the links between the same concepts are traced in the concept trees, then cycles in those trees can be clearly seen. These can represent cognitive processes that very reasonably map across to what the instruction set requires. Some of the links are uni-directional and would be used to get a process started. The others are then cycles - as shown in Figure 2 and there is clearly a sense of order or direction, moving from tree base to leaf and then across to the next tree.

The uni-directional links are as follows:

1. Kitchen, Cupboard and Pot to Pot.
2. Kitchen, Cooker and Hob to Hob.
3. Pot and Water to Water and Tap.

While the process cycles are as follows:

1. Pot and Egg to Egg, Water and Pot.
2. Pot and Heat to Heat and Pot.
3. Pot and Egg to Egg and Heat to Heat and Pot.
4. Hob and Heat to Heat and Hob.

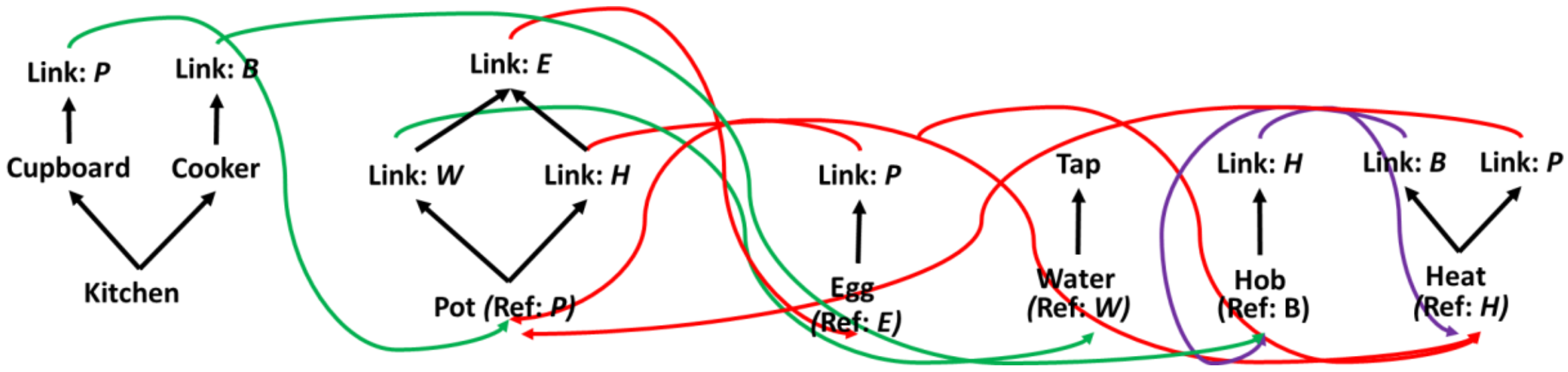

Figure 2. Cognitive Processes.

There is a hint of a process when the concepts are reversed, which is inherent in the construction process and it is more prevalent where the trees are the busiest. The database procedure of normalising through removing duplication in every case is probably not correct in biology, where a concept can be repeated in different trees.

### 7.2 The Ensemble-Hierarchy Structure

This structure has been written about previously [9] and it essentially binds an ensemble with a hierarchy, as a single unit that would be larger than a cortical column in the cortex, for example. It is clear from the work in this paper that the process scripts and construction processes also bind an ensemble with a hierarchy and the following sections describe why that is a good idea. An alternative representation of Figure 2 is also presented that is more biologically-oriented.

### 7.2.1 Robust Construction and the Meaning of Neurons

The ensemble therefore helps to define the outer border for the large concept, but the hierarchy can be more specific with nested or sub-patterns, for example. When creating this structure, one option is to wait for the ensemble to form some links and then copy those links to the hierarchy. The hierarchy therefore, only adds nodes and links when they are created in the ensemble first. This might appear doubtful because the ensemble has no real order to it, but possibly, it is the path that is more important than an exact neuron ordering. When dealing with batch or averaged values for example, the exact order may not be as important. So, for example, the average of ABC is the same as the average of CBA. Therefore, maybe it is the path through the tree structure that matters and the mixed signal through that gets interpreted as the same at the end. This is helpful and allows the construction process some leeway, including timed events through the path. In fact, frequency would determine what nodes are created first. What may matter more for the tree is that it continues to add to what is there, using new branches, for example and does not cycle back, converting the hierarchy into an ensemble.

So, eventually the ensemble will fill-up and the fully-connected nodes will fire each time, when the hierarchy can maintain some level of structure. If processes use the same concept, another question is why not construct long lines of patterns, where some of them repeat the concept when required. Efficiency may be a good-enough reason to share the same concept and the ensemble would force hierarchies to be used over the same node sets. If the ensemble stores abstract representations, then 1 copy of a concept may suffice. If it is an original memory instance however, then repeated nodes would be allowed. When the hierarchy produces structured views from the lower half, it would therefore have to use that abstract instance each time, but this then feeds back to all of the related memory instances. Another advantage is that the hierarchy domain is constrained by what the ensemble represents, although realistically, it must still branch to other places as well. So there are a lot of positive reasons why there should be a close binding between an ensemble structure and a hierarchical one.

### 7.2.2 Example Cognitive Process Network

If the rules of section 6 are added to the hierarchy using the process described in section 7.2.1, then a slightly different network is formed, shown in Figure 3. It is much simpler than the normalised version and cycles can be performed by tracing from any node back through the base 'Pot' node again. Therefore, 'Pot-Water-Egg' is one process and another branch, 'Pot-Heat-Egg' is an alternative process. In this example, 'Egg' has been added to the end of both 'Water' and 'Heat', where those nodes may have already existed, but a real network can also re-wire and possibly place it in-between somewhere. These process cycles are explained in detail in [8].

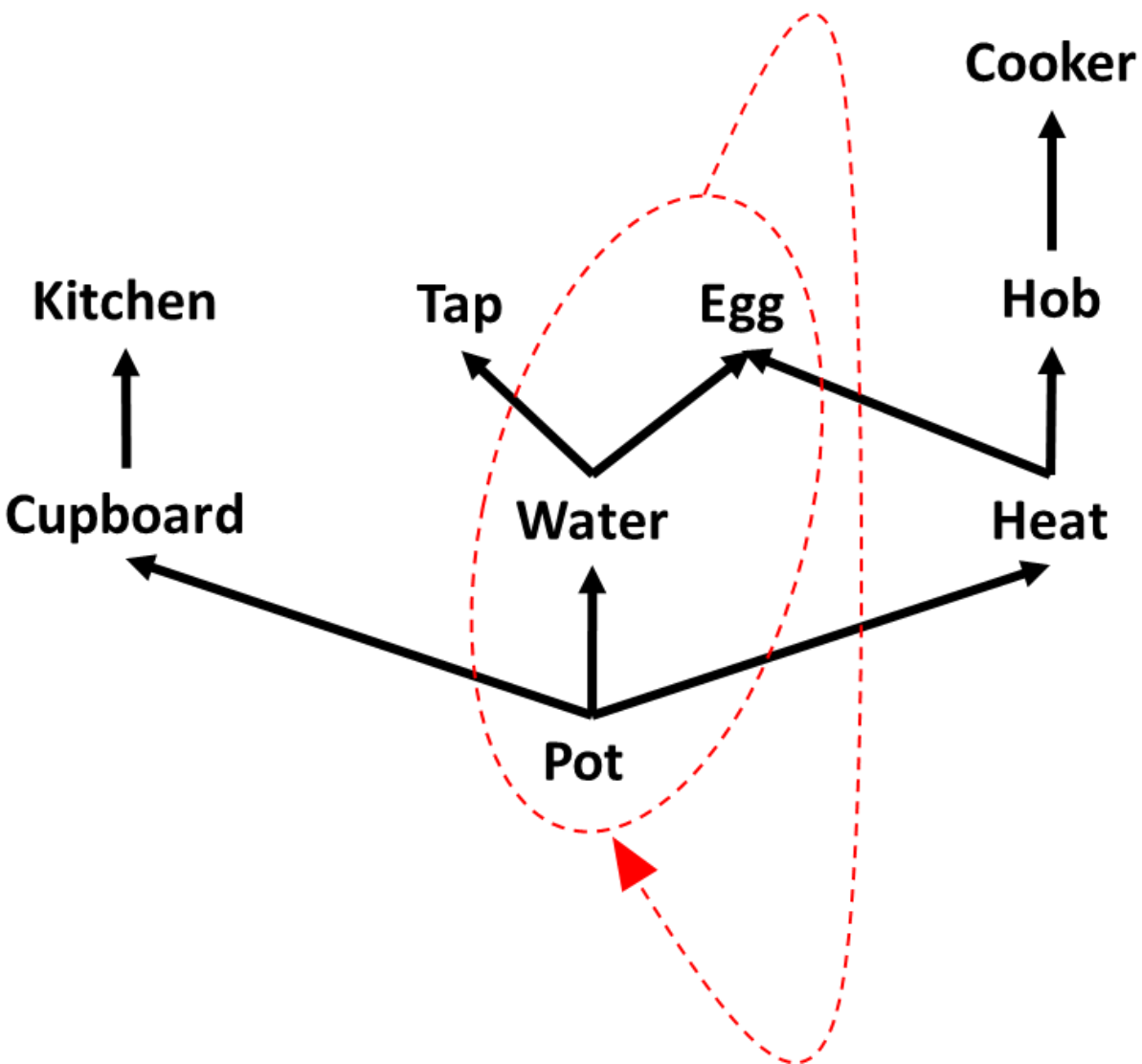

Figure 3. Process Hierarchy without the Normalisation.

## 8 Conclusions

This paper has described a systematic process for searching over distributed information and then making predictions on it. The prediction comes in the form of a related piece of information, found in memory, that is eventually considered significant during the search process. Its realisation is still controlled by the current situation and related memories, and so there are clear constraints as to what may eventually become a prediction. The process is actually one of reduction, to remove the more unlikely possibilities and to direct the search towards the more likely options. The brain-like processes can start by being modelled more like a computer system and later-on, be modelled more as neuronal components. The paper is only suggesting a general architecture and does not give biological specifics. It is known that memories are retrieved when we think about something and obviously, they must be relevant to the situation. Any reasoning process must filter through the information to give a better result and this often involves remembering things not in the current situation. So, while the exact mechanics are not known, the algorithm of this paper does map to general brain processes quite closely and could help to explain how the brain works.

On the biological front, the design fits in well with the cognitive architecture that has been suggested by the author. Converting from ensembles to types has now been suggested more than once, see also [8][15] and it may also assist with the biological function of preparing neurons to fire. For example, a reduced signal from a recognised type to a pattern could help to prepare that pattern for full activation. The different hierarchical structures are put in the context of the cognitive model again, where different types of 'unit of work' have been realised. Building tree structures from nested ensembles is also clear. Neurons that link with each other when they represent the same thing, is now a common feature and can be the basis for a theory about how these regions might link-up.

This paper has also suggested a formal method for describing rules from the pattern ensembles, to produce the cognitive processes. From the formal specification, it is possible to define scenarios in a more centralised way. These can be mapped to an outer model of nested patterns. How each pattern works internally can therefore still be a mystery that requires learning, while the outer framework that they must eventually match with can be properly understood. This largely organic process would allow research to be carried out on less computer hardware, for example. The language also has a mathematical basis, which will help to ensure that all of the rules are consistent with each other and the triple relation gives some direction to each rule. The ensemble-hierarchy structure is now important to the model and further helps to automate the construction processes. Two types of tree can be recognised now, that may or may not be concept trees. The first is the concept tree for memorising image objects and their natural relation to the world. But processes can also be modelled as tree structures and the CPL cognitive language is used to demonstrate this. In fact, the process construction is more horizontal and the memory objects more vertical, but they can both happen in systematic and related ways.

## References

[1] Anderson, J.A., Silverstein, J.W., Ritz, S.A. and Jones, R.A. (1977) Distinctive Features, Categorical Perception, and Probability Learning: Some Applications of a Neural Model, Psychological Review, Vol. 84, No. 5.

[2] Bellman, R. (1957). A Markovian Decision Process. Journal of Mathematics and Mechanics. Vol. 6.

[3] Curbera, F., Goland, Y., Klein, J., Leymann, F., Roller, D., Thatte, S. and Weerawarana, S. (2002). Business Process Execution Language for Web Services BPEL.

[4] Deco, G., Jirsa, V.K., Robinson, P.A., Breakspear, M. and Friston K. (2008). The Dynamic Brain: From Spiking Neurons to Neural Masses and Cortical Fields, PLoS, Comput Biol, Vol. 4, Issue 8, e1000092. doi:10.1371/journal.pcbi.1000092.

[5] Dehaene, S., Changeux, J.P. and Nadal, J.P. (1987). Neural networks that learn temporal sequences by selection. Proceedings of the National Academy of Sciences, Vol. 84, No. 9, pp. 2727 - 2731.

[6] FIPA, http://www.fipa.org/. (last accessed 30/6/17).

[7] Gong, P. and van Leeuwen, C. (2004). Evolution to a small-world network with chaotic units. Europhysics Lett, Vol. 67, pp. 328 - 333.

[8] Greer, K. (2020). New Ideas for Brain Modelling 6, AIMS Biophysics, Vol. 7, Issue 4, pp. 308-322. doi: 10.3934/biophy.2020022.

[9] Greer, K. (2019). New Ideas for Brain Modelling 3, *Cognitive Systems Research*, Vol. 55, pp. 1-13, Elsevier. doi: https://doi.org/10.1016/j.cogsys.2018.12.016.

[10]Greer, K. (2017). New Ideas for Brain Modelling 4, BRAIN. Broad Research in Artificial Intelligence and Neuroscience, Vol. 9, No. 2, pp. 155-167. ISSN 2067-3957.

[11]Greer, K. (2016). A Repeated Signal Difference for Recognising Patterns, BRAIN. Broad Research in Artificial Intelligence and Neuroscience, Vol. 7, No. 3, pp. 139 - 147, ISSN 2067-3957 (online), ISSN 2068 - 0473 (print).

[12]Greer, K. (2014). Concept Trees: Building Dynamic Concepts from Semi-Structured Data using Nature-Inspired Methods, in: Q. Zhu, A.T Azar (eds.), Complex system modelling and control through intelligent soft computations, Studies in Fuzziness and Soft Computing, Springer-Verlag, Germany, Vol. 319, pp. 221 - 252, 2014.

[13]Greer, K. (2012). Turing: Then, Now and Still Key, in: X-S. Yang (eds.), Artificial Intelligence, Evolutionary Computation and Metaheuristics (AIECM) - Turing 2012, Studies in Computational Intelligence, 2013, Vol. 427/2013, pp. 43-62, DOI: 10.1007/978-3-642-29694-9_3, Springer-Verlag Berlin Heidelberg.

[14]Greer, K. (2011). Symbolic Neural Networks for Clustering Higher-Level Concepts, NAUN International Journal of Computers, Issue 3, Vol. 5, pp. 378 - 386, extended version of the WSEAS/EUROPMENT International Conference on Computers and Computing (ICCC'11).

[15]Guigon, E., Grandguillaume, P., Otto, I., Boutkhil, L. and Burnod, Y. (1994). Neural network models of cortical functions based on the computational properties of the cerebral cortex, Journal of Physiology, Vol. 88, pp. 291 - 308, Elsevier.

[16]Hawkins, J. and Ahmad, S. (2016). Why Neurons Have Thousands of Synapses, a Theory of Sequence Memory in Neocortex. Front. Neural Circuits, Vol. 10, No. 23. doi: 10.3389/fncir.2016.00023.

[17]Hawkins, J. and Blakeslee, S. On Intelligence. Times Books, 2004.

[18]Hinton, G.E., Osindero, S. and Teh, Y.-W. (2006). A fast learning algorithm for deep belief nets, Neural computation, Vol. 18, No. 7, pp. 1527 – 1554.

[19]IBM. (2003). An Architectural Blueprint for Autonomic Computing, IBM and Autonomic Computing.

[20]Kandel, E.R. (2001). The Molecular Biology of Memory Storage: A Dialogue Between Genes and Synapses, Science magazine, Vol. 294, No. 5544, pp. 1030 - 1038.

[21]Mastrandreal, R., Gabrielli, A., Piras, F., Spalletta, G., Caldarelli, G. and Gili, T. (2017). Organization and hierarchy of the human functional brain network lead to a chain-like core, Nature Scientific Reports, Vol. 7, No. 4888, DOI:10.1038/s41598-017-04716-3.

[22]Meunier, D., Lambiotte, R. and Bullmore, E.T. (2010). Modular and hierarchically modular organization of brain networks, Frontiers in Neuroscience, Volume 4, Article 200.

[23]Rockstrom, A. and Saracco, R. (1982). SDL-CCITT specification and description language. IEEE Transactions on Communications, Vol. 30, No. 6, pp. 1310 - 1318.

[24]Rubinov, M., Sporns, O., van Leeuwen, C., and Breakspear, M. (2009). Symbiotic relationship between brain structure and dynamics. BMC Neurosci. 10, 55.

[25]Thiagarajan, R.K., Srivastava, A.K., Pujari, A.K. and Bulusu, V.K. (2002). BPML: A Process Modeling Language for Dynamic Business Models, Proceedings of the 4th IEEE Int'l Workshop on Advanced Issues of E-Commerce and Web-Based Information Systems (WECWIS 2002).

[26]Watts DJ and Strogatz SH. (1998). Collective dynamics of 'small world' networks, Nature, Vol. 393, pp. 440 – 442.

[27]Yuste, R. (2011). Dendritic Spines and Distributed Circuits, Neuron, Vol. 71, pp. 772 - 781.